\documentclass[10pt,twocolumn,letterpaper]{article}

\usepackage[pagenumbers]{cvpr} 

\usepackage{graphicx}
\usepackage{subcaption}
\usepackage{amsmath}
\usepackage{amssymb}
\usepackage{booktabs}
\usepackage{pgffor}
\usepackage[export]{adjustbox}
\usepackage[table]{xcolor}
\usepackage{float}
 \usepackage[none]{hyphenat}
 \newcommand{\RomanNumeralCaps}[1]
    {\MakeUppercase{\romannumeral #1}}

\usepackage[pagebackref,breaklinks,colorlinks]{hyperref}

\usepackage[capitalize]{cleveref}
\crefname{section}{Sec.}{Secs.}
\crefname{section}{Section}{Sections}
\crefname{table}{Table}{Tables}
\crefname{table}{Tab.}{Tabs.}

\def\cvprPaperID{11446} 
\def\confName{CVPR}
\def\confYear{2024}

\usepackage[table]{xcolor}
\begin{document}

\title{MiKASA: Multi-Key-Anchor \& Scene-Aware Transformer \\for 3D Visual Grounding}

\author{Chun-Peng Chang, Shaoxiang Wang, Alain Pagani, Didier Stricker\\
DFKI Augmented Vision
}

\maketitle

\begin{abstract}
   3D visual grounding involves matching natural language descriptions with their corresponding objects in 3D spaces. Existing methods often face challenges with accuracy in object recognition and struggle in interpreting complex linguistic queries, particularly with descriptions that involve multiple anchors or are view-dependent. In response, we present the MiKASA (Multi-Key-Anchor Scene-Aware) Transformer. Our novel end-to-end trained model integrates a self-attention-based scene-aware object encoder and an original multi-key-anchor technique, enhancing object recognition accuracy and the understanding of spatial relationships. Furthermore, MiKASA improves the explainability of decision-making, facilitating error diagnosis. Our model achieves the highest overall accuracy in the Referit3D challenge for both the Sr3D and Nr3D datasets, particularly excelling by a large margin in categories that require viewpoint-dependent descriptions. 
   The source code and additional resources for this project are available on GitHub:  https://github.com/dfki-av/MiKASA-3DVG 
\end{abstract}
\section{Introduction}
\label{sec:intro}
3D visual grounding serves as a crucial component in the intersection of natural language processing and computer vision. This task aims to identify and localize objects within a 3D space, using linguistic cues for spatial and semantic grounding. While existing research has made significant strides, challenges remain. Key issues include the lack of explainability in current models, limitations in object recognition within point cloud data, and the complexity of handling intricate spatial relationships.
\begin{figure}[t]
    \centering
    \subcaptionbox{Target category: ``chair" \label{fig:category}}[0.45\linewidth]{
        \includegraphics[width=\linewidth]{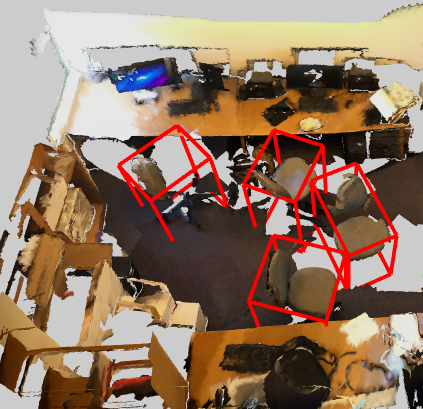}
    }
    \hfill
    \subcaptionbox{``The chair in the front of the blue-lit monitor." \label{fig:cate_sp}}[0.45\linewidth]{
        \includegraphics[width=\linewidth]{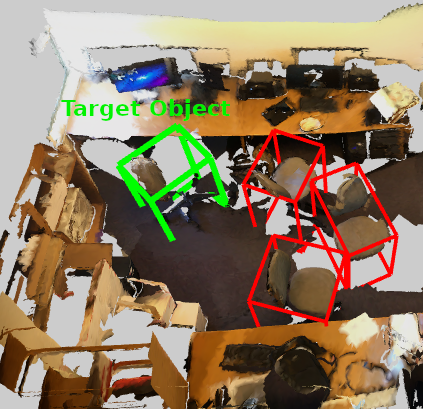}
    }
    \caption{Our methodology utilizes a dual-prediction framework for
    3D visual grounding. First, we assign a target category score 
    based on object categorization, as detailed in \cref{fig:category}. Next, a
    spatial score is integrated according to the object's alignment
    with the textual description, as shown in \cref{fig:cate_sp}.}
    \label{fig:multi_classification}
\end{figure}

Most existing 3D visual grounding models \cite{achlioptas2020, zhao2021,yang2021,jain2021,huang2022} consist of three parts: (1) object encoder, (2) text encoder, and (3) fusion model. The object encoder processes the provided point cloud and generates features in the embedding space. However, because the points in a point cloud are unordered and inconsistent in sparsity \cite{qi2017a}, it is not straightforward to apply the methodology typically used for 2D images. An additional challenge is that 3D point cloud datasets are not as extensive as those for 2D images \cite{guo2020, qi2017}, which makes it difficult for the models to correctly recognize object categories. While enlarging the dataset could conceivably improve performance, we refrain from doing so to ensure a fair comparison with existing state-of-the-art methods. Existing works \cite{qian2022, chen2020a, kim2021} mainly use different techniques such as noise addition, dropping out colors, and transformations to expand the sample space. Though these techniques may increase the stability of the produced object embeddings, the improvement is limited. 

Inspired by previous works\cite{li2014, liu2018, zhang2019} which aims to solve object recognition problem, we leverage the fact that data availability on objects within a specific space can provide valuable insights into the characteristics and relationships of their surrounding entities.
For instance, when we come across a cuboid-shaped object in a kitchen, we may naturally assume that it is a dishwasher. Conversely,
if we spot the same shape in a bathroom, it is more plausible that it is a washing machine. 
Contextual information is crucial in determining the true identity of an object and gives us a nuanced understanding of our surroundings.

Therefore, by incorporating a scene-aware object encoder that considers all nearby objects, we demonstrate that the model's object classification accuracy improves before the data is fed into the fusion model.

Another important task is to represent the spatial
relations between objects. Previous works\cite{huang2022, jain2021} have primarily focused on encoding the absolute locations of objects in world coordinates, a method we find suboptimal. Inspired by how humans use objects as anchors for spatial reasoning and thereby shift the perceptual focus to these anchor objects, our approach hypothesizes that encoding spatial relations relative to anchors enhances the model's ability to accurately identify objects. This hypothesis has been empirically validated in our experiments.

Therefore, we introduce the multi-key-anchor concept to enhance spatial understanding in 3D models. This approach translates the coordinates of potential anchors relative to a target object and explicitly evaluates the importance of nearby objects based on textual descriptions. Notably, models like PointNet++\cite{qi2017a} are designed for rotational invariance, often leading to directional ambiguity in object features. Our method mitigates this issue by leveraging the spatial context of key nearby objects, thereby implicitly suggesting the orientation of the target object. For example, a chair is typically placed facing a table or against a wall, and the presence of the table or the wall implicitly defines the direction of the chair. This approach provides a nuanced understanding of object orientation and spatial placement. We validate this claim experimentally, by showing a higher improvement in view-dependent cases compared to the overall accuracy.

Finally, instead of treating the information as input for a black-box fusion module, we introduce a new architecture that employs a more novel approach to the 3D visual grounding task. Inspired by human-like object searching behavior—for instance, when given the instruction ``The chair in the front of the blue-lit monitor.", one would first identify all the chairs in the room before pinpointing the specific target, as illustrated in \cref{fig:multi_classification}. Our model employs late fusion and generates two distinct output scores. The first score aims to identify the target object category, while the second assesses the location and language expression, informed by spatial data. These scores are designed to collaborate, mitigating the influence of objects that may superficially resemble the target or occupy positions that seem to fit the verbal description but are not the intended objects. We fuse these scores through a strategic fusion mechanism, which enables the final target to distinguish itself more clearly from distractors. By examining the two scores and the final result, one can better diagnose the types of errors the model may make, making the decision process more explainable.\\

The main contributions of our work can be summarized as follows:
\begin{itemize}
    \item We introduce a scene-aware object encoder that considers the contextual information and increases models ability to understand the object category. 
    \item We present the multi-key-anchor technique, which enhances spatial understanding. This approach redefines coordinates relative to target objects and explicitly assesses the importance of nearby objects through textual context. It addresses the directional ambiguity often found in rotationally invariant models like PointNet++\cite{qi2017a}, by using spatial contexts to imply target object orientation.
    \item We develop a novel, end-to-end trainable and explainable architecture, that leverages late fusion to separately process distinct aspects of the data, thereby enhancing the model's accuracy and explainability. 
\end{itemize}

\begin{figure*}[t] 
\centering 
\includegraphics[width=\linewidth]{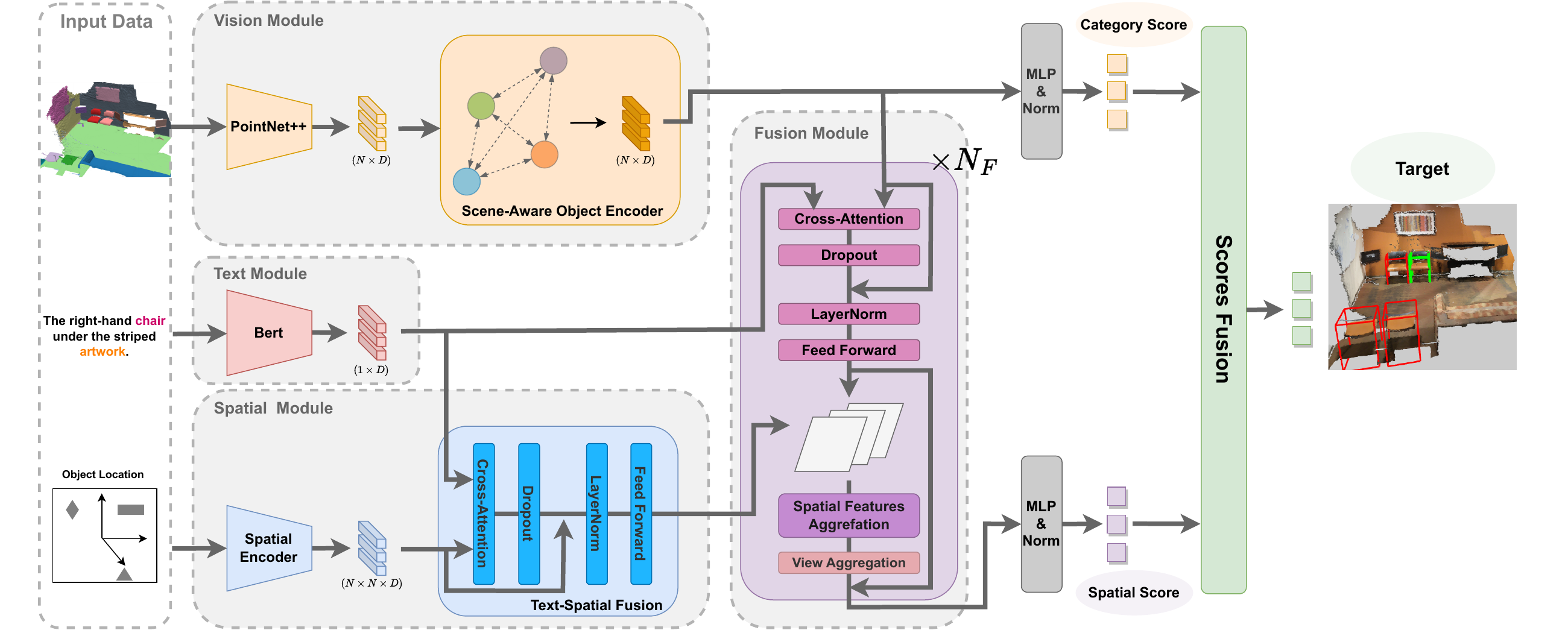} 
\caption{Architecture of our 3D Visual Grounding Model, which includes four main modules: a text encoder (Bert), a vision module with a scene-aware object encoder, a spatial module that fuses spatial and textual data, and a multi-layered fusion module. The fusion module combines text, spatial, and object features, employing a dual-scoring system for enhanced object category identification and spatial-language assessment.} 
\label{fig:architecture} 
\end{figure*}
\section{Related Works}
\label{sec:related_work}
\subsection{3D Visual Grounding}
3D visual grounding, intersecting computer vision and natural language processing, focuses on identifying objects in 3D spaces using language. Unlike 2D grounding that relies on images, 3D visual grounding utilizes point cloud data. This transition from 2D models to point clouds introduces a new layer of complexity, and marks a distinct shift from conventional 2D grounding models \cite{kazemzadeh2014, plummer2016, yu2016} due to the unique nature of point clouds\cite{guo2020}. Our experiments and comparative analyses are conducted using the Referit3D benchmark\cite{achlioptas2020}. 

The methods for feature fusion for 3D visual grounding have recently evolved. Initially, graph-based algorithms were predominantly used\cite{achlioptas2020, huang2021}. However, with the rise of transformer models, the focus has shifted towards these, given their effectiveness in multimodal data fusion\cite{abdelreheem2022, he2021, zhao2021, yang2021, jain2021, huang2022}. Notable among these are LAR\cite{bakr2022}, which synthesizes 2D clues from point clouds for 3D visual grounding, SAT\cite{yang2021} that leverages 2D image semantics in training, and 3DVG-Transformer\cite{zhao2021}, using a relation-aware approach with contextual clues for proposal generation. To enhance point cloud data representation, MVT\cite{huang2022} maps 3D scenes into multi-view spaces, aggregating positional information from various perspectives. ViewRefer\cite{guo23} builds upon this by resolving view discrepancies through the integration of multi-view inputs and inter-view attention. Furthermore, ViewRefer utilizes GPT-3\cite{brown20} to generate multiple geometry-consistent descriptions from a single grounding text, thereby enriching the model's interpretation of 3D environments.

\subsection{Context-Aware Object Recognition}
3D object recognition and segmentation using point clouds is a foundational task in computer vision. Most established methodologies \cite{qi2017, qi2017a} predominantly utilize appearance features such as color and position to define objects, achieving impressive outcomes. However, in contexts where a comprehensive scene is involved, such as a room filled with diverse objects, methods that incorporate inter-object relationships and spatial context \cite{galleguillos2010, zhang2019,li2014, liu2018, teney2016, yao2012} offer better performance. These models do not just consider individual object features, but also the relative placements and attributes of neighboring objects. This allows for a more sophisticated understanding of complex 3D spaces.

\subsection{Multi-Modal Features Fusion}
In multi-modal features fusion, two primary fusion approaches exist: early and late. Early fusion merges features from different modalities at the outset and trains a unified model, as commonly seen in 3D visual grounding work\cite{achlioptas2020,zhao2021,yang2021,jain2021,huang2022}. This approach benefits from direct inter-modality interactions but can be challenging to fine-tune and lacks transparency regarding its decision-making process. Late fusion, on the other hand, processes each modality separately and fuses the resulting logits or decision scores. Various techniques, from simple averaging to attention-based methods, are used in existing works\cite{zhang2021, madichetty2021, zhang2020, pranesh2022, alam2018,xue2021}. The fusion strategy significantly impacts the system's robustness and accuracy, especially when modalities provide conflicting cues.

\section{Method}
\label{sec:method}
\cref{fig:architecture} presents our novel architecture for the 3D visual grounding task, including key modules: a vision module, a text encoder (Bert\cite{devlin2018}), a spatial module, and a fusion module. We maintain configurations for Bert and PointNet++\cite{qi2017a} as in MVT\cite{huang2022} for consistent comparison. The vision module, with our scene-aware object encoder, refines object features by considering surrounding objects. The spatial module encodes and merges spatial features with textual data from Bert using a transformer encoder. In the fusion module, comprising \(N_F\) layers, object-spatial features are progressively refined with text and spatial information, enhancing spatial features through refined anchor information.
Our model concludes with a dual-scoring system, generating scores for object category identification and spatial feature assessment. This approach mitigates the influence of distractors and enhances the explainability of the model's decisions.
\begin{figure*}[t] 
\centering 
\includegraphics[width=\linewidth]{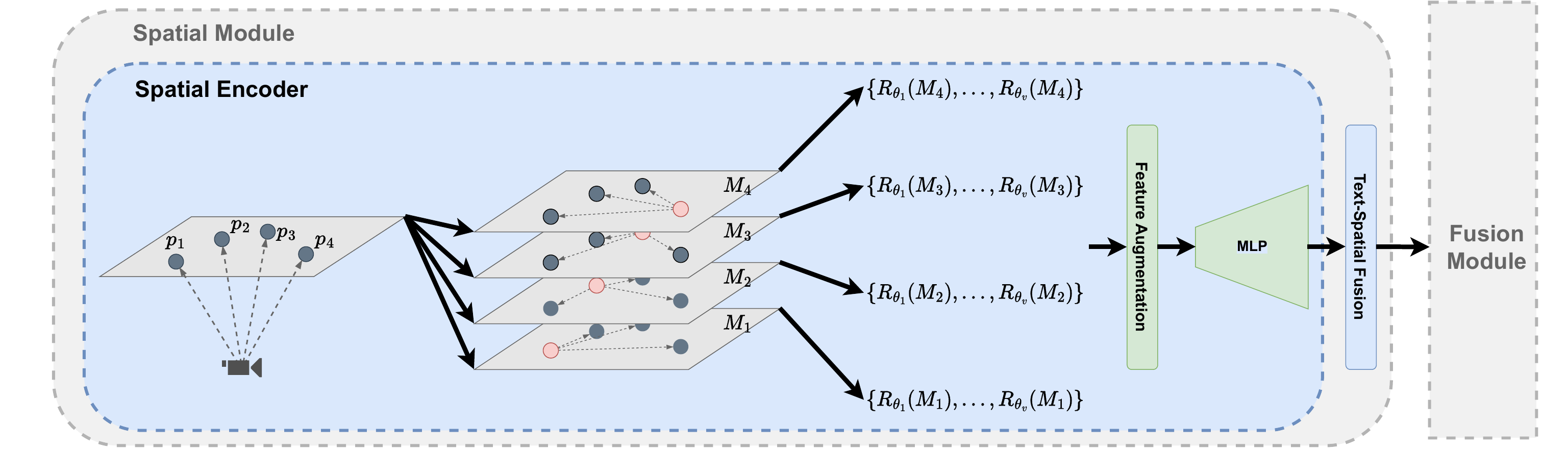} 
\caption{Our novel spatial module captures relative spatial information from a single viewpoint by treating each object in the scene as a potential anchor. This approach generates unique spatial maps, each offering a different perspective of the scene. These maps are then undergo feature augmentation, where distances and angles are calculated, followed by normalization and scaling. Subsequently, a MLP layer is employed to transform these low-dimensional features into higher-dimensional ones for effective fusion with textual data.} 
\label{fig:spatial_map} 
\end{figure*}

\subsection{Data Augmentation}
\label{sec:data_augmentation}
Given the limitations and scarcity of 3D point cloud datasets, data augmentation emerges as a crucial strategy. To improve the model against overfitting and enhance its ability to generalize essential features, we've adopted different data augmentation techniques. This includes the notable multi-view augmentation as presented in MVT\cite{huang2022}, which has demonstrated effectiveness. Additionally, we place emphasis on augmenting color features, adjusting contrast, and introducing noise. More details can be found in the supplementary materials.

\subsection{Scene-Aware Object Encoder}
Traditional point cloud methods usually analyze objects separately, missing contextual clues from other objects in the same room. To overcome this, our model includes a scene-aware object encoder that gathers information from surrounding objects. This approach overcomes the limits of standard object encoders. By considering this additional context, our model better understands object categories, leading to improved accuracy and performance.

While graph-based algorithms such as DGCNN\cite{wang2019}, GCN\cite{kipf2017}, and GAT\cite{velickovic2018} are often used for similar tasks, they present their own set of challenges, such as computational intensity and the complexity of defining a suitable distance metric for determining neighbors. As a result, we opted for a self-attention mechanism for feature aggregation. This choice offers a range of advantages, including computational efficiency, ease of training, and consistently superior performance, making it the most fitting solution for the objectives of our model. For \(N\) objects in the room, represented as \(O = \{O_1, O_2, \ldots, O_N \}\) where each \(O_i\) is a feature vector with \(D\) dimensions, the scene-aware object features \(O^{sa}_i\) are computed using self-attention as defined in \cref{eq:selfattnobj}. In this process, \(Q = W_Q \cdot O\), \(K = W_K \cdot O\), and \(V = W_V \cdot O\) represent the queries, keys, and values, respectively, each transformed by their respective learnable weight matrices \(W_Q\), \(W_K\), and \(W_V\)\cite{vaswani2017}. By aggregating this information, particularly through the weighted sum of values (\(V\)), each object feature \(O^{sa}_i\) becomes enriched with contextual data from its surroundings. This approach ensures that the object features capture more than just individual properties.

\begin{equation}
O^{sa}_{i} = \sum_{j=1}^{N} \frac{\exp(Q_{i} \cdot K_{j})}{\sum_{k=1}^{N} \exp(Q_{i} \cdot K_{k})} \cdot V_{j}
\label{eq:selfattnobj}
\end{equation}

\subsection{Spatial Features Encoding}
\label{sec:SREMA}
Instead of relying on a single viewpoint for understanding a 3D space, our model employs a novel multi-anchor strategy to better comprehend spatial relationships, as shown in \cref{fig:spatial_map}. For \(N\) objects in the room, this approach involves generating \(N\) feature maps, each representing relative positions from \(N\) unique local coordinate systems. We define a set of spatial maps for a given set of object coordinates, \(P = \{p_1, p_2, \ldots, p_N\}\), as \(M = \{M_1, M_2, \ldots, M_N\}\). Each map \(M_i\), defined in \cref{eq:Mi}, is composed of the relative positions of all other objects \(A = \{a_j  | a_j\in P \text{ and } j \neq i\}\) to a target object \(p_i\).
\begin{equation}
M_i = \{(a_j-p_i) | a_j \in A\}
\label{eq:Mi}
\end{equation}

To enhance robustness against varying initial viewpoints, our model incorporates the viewpoint augmentation strategy similar to MVT\cite{huang2022}. We utilize a rotation matrix \( R_{\theta_k}\) for each viewpoint, which rotates the entire map by an angle \( \theta_k \). This approach results in an augmented set \( M^{aug} \), defined in \cref{eq:m_aug}, wherein each map \( M_i \) from the original set \( M \) is represented under \( v \) different rotated views. 
\begin{equation}
M^{aug} = \{R_{\theta_k}(M_i) \ | \ M_i \in M, \ k \in \{1, 2, \ldots, v\}\}
\label{eq:m_aug}
\end{equation}

Each element of a map then goes through feature augmentation to incorporate additional spatial features, including distance and angles. These augmented features are later on normalized and scaled to ensure stability. Normalization and scaling details are in the supplementary materials.

Subsequently, the map features undergo a dimensional transformation \(\mathcal{T}\) to align with the dimension \(D\), as described in \cref{eq:trans_to_D}. This process prepares the map features for effective fusion with other features at later stages of the model.
\begin{equation}
M^D = \mathcal{T}(M^{aug}, D)
\label{eq:trans_to_D}
\end{equation}

\subsection{Multi-Modal Feature Fusion}
To create a multi-anchor spatial map, we combine features from various modalities. Specifically, we merge the object feature, spatial feature, and the text feature. This integrative approach generates a new detailed spatial feature that accurately represents an object's location within a room.
\subsubsection{Text-Spatial Fusion}
To optimize computational efficiency, we merge text and spatial features at an initial stage instead of in the multi-layer fusion module. We use a single cross-attention layer, represented by \( \mathcal{A} \), followed by a subsequent feedforward layer, denoted as \( \mathcal{F} \). This approach is employed instead of merging these features within the fusion module alongside additional features. With \(N\) spatial maps at our disposal, this method provides a computationally economical means of feature integration.
The fusion of the spatial map \(M\) with the textual information \(T\) synchronizes spatial features with corresponding text data, as expressed in \cref{eq:map}:
\begin{equation}
M^{text}_i = \mathcal{F} (\mathcal{A}(M_i, T)), \ \forall i \in \{1, 2, \ldots, N\}
\label{eq:map}
\end{equation}

\subsubsection{\bf Fusion Module}
Our fusion module's architecture consists of four key components in each layer: (1) Text-Object Fusion, (2) Object-Spatial Fusion, (3) Spatial Feature Aggregation, and (4) View Aggregation. As data progresses through these layers, the model incrementally adjusts anchor weightings and extracts relevant spatial information based on textual input. This progressive fusion method refines features within specific spatial maps while dynamically updating anchor features across the maps.

The effectiveness of this approach is particularly notable in complex scenarios. Take, for example, the instruction ``Choose the suitcase that is in front of the bed near the window curtains." Here, the primary anchor (``bed") is contextualized by another reference point (``curtains"), underscoring the need for coherent interaction between different spatial maps. This scenario highlights the value of our fusion process in managing intricate spatial relations. \\

\noindent{\bf Text-Object fusion \&  Object-Spatial Fusion:}\\
The Text-Object Fusion component employs an architectural approach akin to that of the Text-Spatial Fusion in \cref{eq:map}, incorporating both a cross-attention layer and a feedforward neural network layer. Subsequently, the text-object features are integrated into each spatial map by addition and linear transformation. The augmented map now encapsulates not only the spatial information but also the characteristics of the anchor objects. 

\begin{figure}[t]
  \centering
   \includegraphics[width=0.95\linewidth]{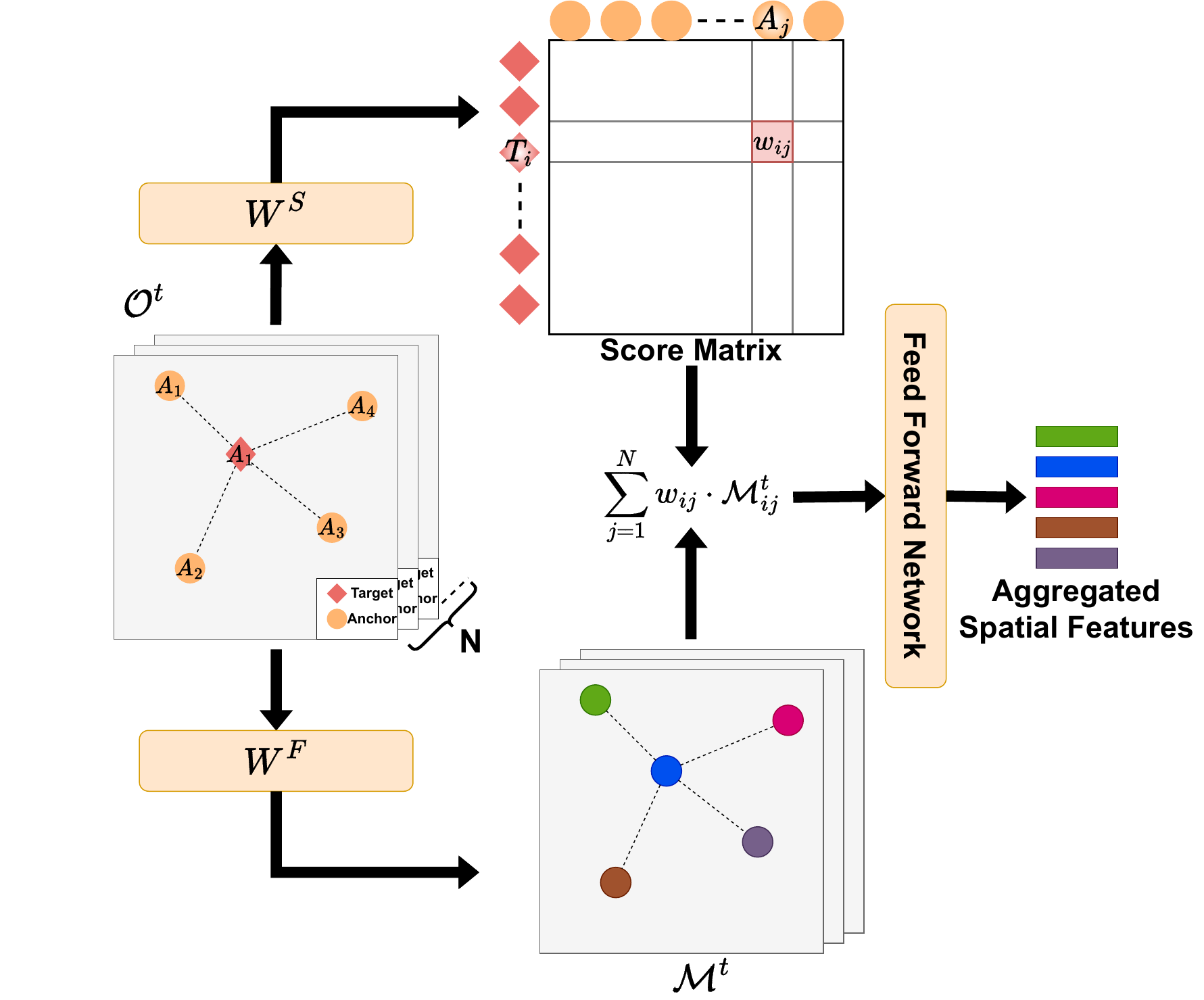}

   \caption{Our novel attention-based spatial feature aggregation. Each map designates a different object as the target, while treating all other objects as anchors. The importance of each anchor relative to the potential target object is represented in row \(i\) of the score matrix, indicating the relevance of each anchor in the context of the target, where \(W_S\) and \(W_F\) are learnable weight matrices. }
   \label{fig:multi_anchor_agg}
\end{figure}
\noindent{\bf Spatial Feature Aggregation \& View Aggregation:}\\
In our model, essential information is gathered from the fused feature maps through an attention-based aggregation stage, as depicted in \cref{fig:multi_anchor_agg}. This stage employs attention mechanisms to determine the relevance of potential key anchors relative to the target object. The process effectively consolidates these weighted anchors into a single spatial feature for each object, enhancing the model's grasp of the spatial relationships of each object within the scene. Subsequently, to ensure robustness against the initial viewpoint, the features undergo further refinement by aggregating views from various perspectives, a method elaborated in \cref{sec:data_augmentation}.\\

\noindent{\bf Progressive Feature Enhancement:}\\
The fusion module's effectiveness hinges on the iterative refinement of object-spatial features \( \mathcal{O}^t \), as detailed in \cref{eq:sp_aggr}. In this process, \( w^i_j \) denotes the significance of anchor \( j \) within the fused feature map \( \mathcal{M}^t_i \), as illustrated in \cref{fig:multi_anchor_agg}. This approach integrates aggregated spatial features into object features, facilitating progressive enhancement across spatial maps. As a result, both anchor and spatial features undergo continuous refinement with each layer, progressively improving the model's ability to accurately represent the spatial dynamics of the scene.
\begin{equation}
\mathcal{O}^{t+1}_i = \mathcal{O}^t_i + \sum_{j=1}^{N} w_{ij} \cdot \mathcal{M}^t_{ij}, \quad \forall i \in \{1, 2, \ldots, N\}
\label{eq:sp_aggr}
\end{equation}

\subsection{Multi-Modal Predictions Fusion}
Previous works \cite{achlioptas2020, yang2021, huang2022} treat the fusion process as a ``black box", where information is aggregated in the early stages and then passed to a fusion module such as a transformer decoder to produce a single set of logits. While this method is effective, it offers limited insight into the intricacies of the fusion process. In contrast, our model adopts a segmented approach to the grounding problem, tackling it through two distinct subtasks. The first subtask determines the target category score by evaluating the object’s correspondence with the text-described target category. Subsequently, a spatial score is computed to assess how well the object's spatial configuration aligns with the spatial description provided in the text.

In our methodology, directly merging logits, \(f_1(X; \theta_1)\) and \(f_2(X; \theta_2)\) from different modalities can lead to suboptimal results due to discrepancies in their scales. To mitigate this, we employ a normalization function, \( g \) to transform each logit to have zero mean and unit variance, facilitating a uniform scale. Subsequently, we assign weights, denoted as $\lambda$ and $\mu$, which are optimized as hyperparameters through experimentation, detailed in the supplementary materials. These weights are applied to the normalized logits, enabling a balanced integration of the modalities. The resulting final prediction \(\mathcal{P}\) is represented as in \cref{eq:logits_fusion}.

\begin{equation}
\mathcal{P} = \lambda \cdot g(f_1(X; \theta_1)) + \mu \cdot g(f_2(X; \theta_2))
\label{eq:logits_fusion}
\end{equation}

\subsection{Loss Function}
The loss function \( \mathcal{L} \) employed in our model is a composite of multiple terms, each aimed at a specific aspect of the 3D visual grounding task. Formally, the loss is defined as \cref{eq:loss_func}.
\begin{equation}
\mathcal{L} = L_{\text{ref}} + \alpha L_{\text{text}} + \beta L_{\text{obj}} + \gamma L_{\text{obj\_scene}}
\label{eq:loss_func}
\end{equation}
where all terms are computed using cross-entropy loss. \( L_{\text{ref}} \) is the primary loss that evaluates the reference results, measuring how accurately the model identifies the correct target among distractors. The auxiliary losses serve to fine-tune different components of the model. Specifically, \( L_{\text{text}} \) evaluates the model's ability to correctly categorize the target object from a given sentence (e.g., recognizing ``table" as the target category in the sentence ``a table near the window"). \( L_{\text{obj}} \) evaluates the categorization of all objects in the 3D space. \( L_{\text{obj\_scene}} \), which we might alternatively name ``Scene-Aware Object Categorization Loss", assesses the categorization performance but does so after the scene-aware object encoder has been applied. The hyperparameters \( \alpha \), \( \beta \), and \( \gamma \) control the relative contributions of these auxiliary losses and the choice of their value is discussed in the supplementary material.

\section{Experiment}
\definecolor{mydarkgreen}{RGB}{0,80,0}
\definecolor{mydarkred}{RGB}{200,0,0}

\begin{table*}[t]
\setlength{\tabcolsep}{4.4pt}
\centering
\small{
\begin{tabular}{l | ccccc | ccccc} 
\hline
\rowcolor[gray]{0.95}
 Model & \multicolumn{5}{c|}{Sr3D} & \multicolumn{5}{c|}{Nr3D} \\
\hline
 & Overall & Easy & Hard & VD\textcolor{red}{\textsuperscript{$\ast$}} & VI\textcolor{blue}{\textsuperscript{$\dagger$}} & Overall & Easy & Hard & VD\textcolor{red}{\textsuperscript{$\ast$}} & VI\textcolor{blue}{\textsuperscript{$\dagger$}} \\ [0.5ex]
\hline
ReferIt3D\cite{achlioptas2020}\quad\scriptsize{\textit{ECCV 20}}& 40.8\% & 44.7\% & 31.5\% & 39.2\% & 40.8\%& 35.6\% & 43.6\% & 27.9\% & 32.5\% & 37.1\%  \\ 
InstanceRefer\cite{yuan2021}\quad\scriptsize{\textit{ICCV 21}}& 48.0\% & 51.1\% & 40.5\% & 45.4\% & 48.1\% & 38.8\% & 46.0\% & 31.8\% & 34.5\% & 41.9\%  \\ 
3DVG-Transf.\cite{zhao2021}\quad\scriptsize{\textit{ICCV 21}} & 51.4\% & 54.2\% & 44.9\% & 44.6\% & 51.7\% & 40.8\% & 48.5\% & 34.8\% & 34.8\% & 43.7\%  \\ 
SAT\cite{yang2021}\quad\scriptsize{\textit{ICCV 21}}& 57.9\% & 61.2\% & 50.0\% & 49.2\% & 58.3\% & 49.2\% & 56.3\% & 42.4\% & 46.9\% & 50.4\% \\ 
MVT\cite{huang2022}\quad\scriptsize{\textit{CVPR 22}}& 64.5\% & 66.9\% & 58.8\% & 58.4\% & 64.7\% & 55.1\% & 61.3\% & 49.1\% & 54.3\% & 55.4\%  \\ 
BUTD-DETR\cite{jain2021}\quad\scriptsize{\textit{ECCV 22}}& 67.0\% & 68.6\%&63.2\% & 53.0\% & 67.6\%& 54.6\% & 60.7\% & 48.4\% & 46.0\% & 58.0\%  \\ 
NS3D\cite{hsu2023}\quad\scriptsize{\textit{CVPR 23}}& 62.7\% & 64.0\% & 59.6\% & 62.0\% & 62.7\% & - & - & - & - & -  \\
ViewRefer\cite{guo23}\quad\scriptsize{\textit{ICCV 23}}& 67.0\% & 68.9\% & 62.1\% & 52.2\% & 67.7\% & 56.0\% & 63.0\% & 49.7\% & 55.1\% & 56.8\%  \\ 
\bottomrule
Ours             & {\bf 75.2\%} & {\bf 78.6\%} & {\bf 67.3}\% & {\bf 70.4\%} & {\bf 75.4\%} & {\bf 64.4\%} & {\bf 69.7\%} & {\bf 59.4\%} & {\bf 65.4\%} & {\bf 64.0\%}  \\
vs. BUTD-DETR\cite{jain2021}\quad\scriptsize{\textit{ECCV 22}} & \color{mydarkgreen} +8.2\% & \color{mydarkgreen} +10.0\% & \color{mydarkgreen} +4.1\% & \color{mydarkgreen} {\bf +17.4\%} & \color{mydarkgreen} +7.8\% & \color{mydarkgreen} +9.8\% & \color{mydarkgreen} +9.0\% & \color{mydarkgreen} +11.0\% & \color{mydarkgreen} \bf+19.4\% & \color{mydarkgreen} +6.0\%  \\
vs. Ns3D\cite{hsu2023}\quad\scriptsize{\textit{CVPR 23}} & \color{mydarkgreen} +12.5\% & \color{mydarkgreen} +14.6\% & \color{mydarkgreen} +7.7\% & \color{mydarkgreen} {\bf +8.4\%} & \color{mydarkgreen} +12.7\%& - & - & - & - & -  \\
vs. ViewRefer\cite{guo23}\quad\scriptsize{\textit{ICCV 23}} & \color{mydarkgreen} +8.2\% & \color{mydarkgreen} +9.7\% & \color{mydarkgreen} +5.2\% & \color{mydarkgreen} {\bf +18.2\%} & \color{mydarkgreen} +7.7\% & \color{mydarkgreen} +8.4\% & \color{mydarkgreen} +6.7\% & \color{mydarkgreen} +9.7\% & \color{mydarkgreen} {\bf +10.3\%} & \color{mydarkgreen} +7.2\%  \\
\hline
\end{tabular}
}

\caption{Comparative analysis of the accuracy of our end-to-end solution against others on the Sr3D and Nr3D Challenges, highlighting MiKASA's enhancements. Notably, in the view-dependent category, our solution demonstrates exceptional performance improvements, underscoring MiKASA's superior accuracy and effectiveness over previous methods. \\   \textcolor{red}{\textsuperscript{$\ast$}}\footnotesize{View-Dependent}, \textcolor{blue}{\textsuperscript{$\dagger$}}\footnotesize{View-Independent}}

\label{table:baseline_comp}
\end{table*}

\subsection{Datasets}
\noindent{\bf Natural Reference in 3D (Nr3D) \cite{achlioptas2020} :}\\
The dataset consists of 41.5k human utterances describing 707 unique 3D indoor scenes from ScanNet \cite{dai2017}. It contains 76 fine-grained object classes. These utterances were collected during a reference game played by two individuals. One person acted as the speaker, selecting an object from a set of distractors (ranging from 1 to 6), while the other person identified the target object based on the speaker's instructions. Each scenario in the data can be categorized as easy/hard and view-dependent/view-independent, depending on the number of distractors and whether the utterances depend on a specific viewpoint.\\

\noindent{\bf Spatial Reference in 3D (Sr3D/Sr3D+) \cite{achlioptas2020} :}\\
Similar to Nr3D, Sr3D dataset contains 83.5k synthetic utterances describing the 3D indoor scenes from ScanNet \cite{dai2017}. Each utterance was generated from the template \[target\_class + spatial\_relation + anchor\_class(es)\]
Sr3D+ expand the dataset by sampling that did not
contain more than one distractor were added to Sr3D . This resulted in an increase to 114.5k total
utterances.

\subsection{Experimental Setup}
\subsubsection{Implementation Details}
In the implementation of our proposed architecture, we utilize a pre-trained BERT\cite{devlin2018} model as the text encoder, generating a 768-dimensional output. The object encoder is implemented using the PointNet++\cite{qi2017a} framework. To ensure a fair comparison with the MVT\cite{huang2022}, the settings for both the text and object encoders are aligned with those used in MVT. Our fusion module is composed of three layers. Importantly, all modules—including the text encoder, object encoder, and fusion module—are trained end-to-end, negating the need to train each component separately. The optimization is carried out using the Adam optimizer with a batch size of 12. All experiments were conducted on an A100 GPU.
\subsubsection{Evaluation Metrics}
In our experiments, we focus on the datasets from Referit3D, namely Nr3D and Sr3D. Evaluation proposals are generated directly from the ground truth annotations. The primary metric for evaluation is accuracy, which gauges the model's ability to successfully identify the correct target among various distractors. A successful match is defined as the model accurately pointing out the designated target from a pool of distractors in the 3D space.

\begin{table*}[t]
\centering
\begin{tabular}{l | ccccc} 
\hline
\rowcolor[gray]{0.95}
Component &  Overall&Easy&Hard&VD&VI\\
\hline
w/o Spatial encoder & \(45.0\%\pm 0.4\%\) & \(53.4\%\pm 0.6\%\) & \(36.9\%\pm 0.6\%\) & \(43.4\%\pm 0.4\%\) & \(45.8\%\pm 0.6\%\)\\ 

w/o Feedforward layer & \(62.4\%\pm 0.2\%\)& \(69.0\%\pm 0.2\%\) & \(56.1\%\pm 0.6\%\) & \(62.4\%\pm 0.2\%\) & \(62.4\%\pm 0.2\%\) \\ 
w/o Text-Spatial fusion & \(62.6\%\pm 0.1\%\)& \(69.0\%\pm 0.3\%\) & \(56.4\%\pm 0.2\%\) & \(61.8\%\pm 0.6\%\) & \(62.9\%\pm 0.1\%\) \\  
\hline
\end{tabular}
\caption{Ablations of the spatial module on Nr3D, results highlighting the essentiality of each component.}
\label{table:spatialinfo}
\end{table*}

\subsubsection{Baseline Comparison}
\Cref{table:baseline_comp} provides a comparative analysis of MiKASA against existing models in the Sr3D and Nr3D challenges\cite{achlioptas2020}. MiKASA leads in overall accuracy for both Sr3D and Nr3D, achieving 75.2\% and 64.4\% respectively. Specifically, it demonstrates exceptional performance in the view-dependent category, achieving 70.4\% in Sr3D and 65.4\% in Nr3D, and significantly surpasses previous works. This performance underlines its capability to handle complex scenarios requiring changes in viewpoint, proving the effectiveness of our multi-key-anchor and features fusion strategy.
\definecolor{mydarkblue}{RGB}{120,50,255}
\begin{figure*}[t]
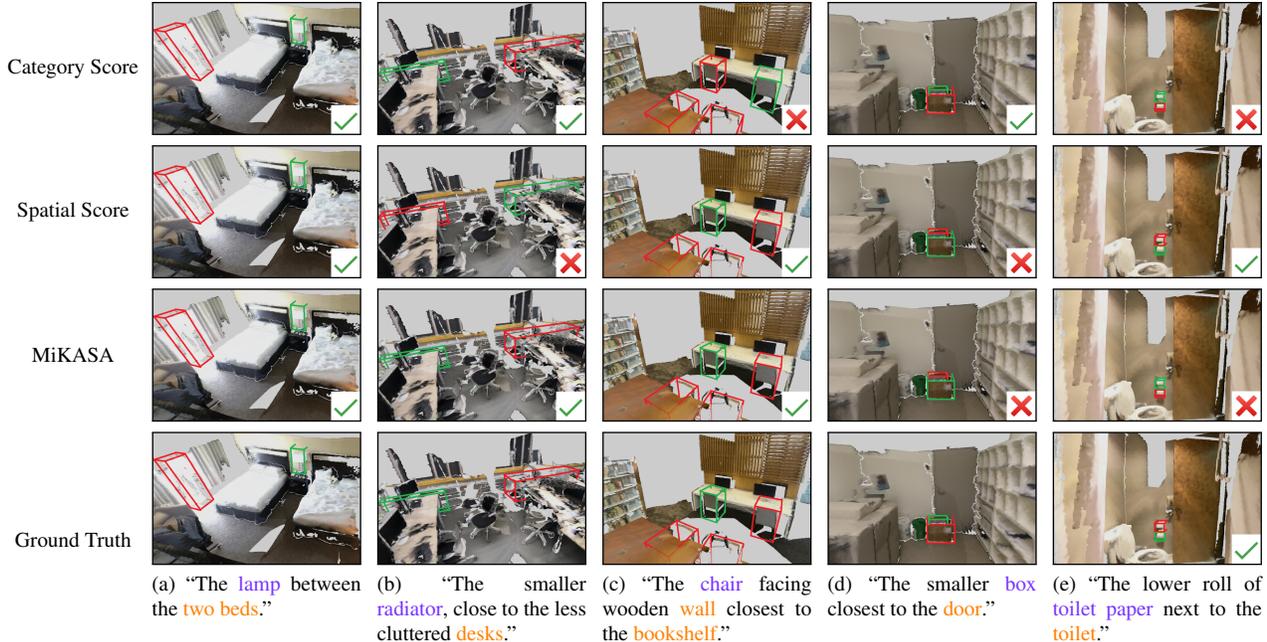

\captionsetup{skip=3pt}
    \centering
    \begin{minipage}{0.11\textwidth}
        \centering
        \footnotesize Category Score
    \end{minipage}
    \vspace*{0.1cm}
    \begin{minipage}{0.85\textwidth}
        \centering
        \foreach \i/\framecolor in{
    1/green,
    2/green,
    3/red,
    4/green,
    5/red} {            {\includegraphics[width=0.185\textwidth, cfbox={black} 0.5pt 0pt]{figures/vis/score1_\i.png}}
            \ifnum\i<5
                \hfill
            \fi
        }
    \end{minipage}    
    \vspace*{0.1cm}
    \begin{minipage}{0.11\textwidth}
        \centering
        \footnotesize Spatial Score
    \end{minipage}
    \begin{minipage}{0.85\textwidth}
        \centering
        \foreach \i/\framecolor in{
    1/green,
    2/red,
    3/green,
    4/red,
    5/green} {            {\includegraphics[width=0.185\textwidth, cfbox={black} 0.5pt 0pt]{figures/vis/score2_\i.png}}
            \ifnum\i<5
                \hfill
            \fi
        }
    \end{minipage}
    \vspace*{0.1cm}
    \begin{minipage}{0.11\textwidth}
        \centering
        \footnotesize MiKASA
    \end{minipage}
    \begin{minipage}{0.85\textwidth}
        \centering
        \foreach \i/\framecolor in{
    1/green,
    2/green,
    3/green,
    4/red,
    5/red} {            {\includegraphics[width=0.185\textwidth, cfbox={black} 0.5pt 0pt]{figures/vis/finalchoice_\i.png}}
            \ifnum\i<5
                \hfill
            \fi
        }
    \end{minipage}

    \begin{minipage}{0.11\textwidth}
        \centering
        \footnotesize Ground Truth
    \end{minipage}
    \begin{minipage}{0.85\textwidth}
        \centering
        \foreach \i/\captiontext in {
    1/{``The \color{mydarkblue}lamp \color{black}between the \color{orange}two beds\color{black}."},
    2/{``The smaller \color{mydarkblue}radiator\color{black}, close to the less cluttered \color{orange}desks\color{black}."},
    3/{``The \color{mydarkblue}chair \color{black}facing wooden \color{orange}wall \color{black}closest to the \color{orange}bookshelf\color{black}."},
    4/{``The smaller \color{mydarkblue}box \color{black}closest to the \color{orange}door\color{black}."},
    5/{``The lower roll of \color{mydarkblue}toilet paper \color{black}next to the \color{orange}toilet\color{black}."}} {            \expandafter\subcaptionbox\expandafter{\captiontext}{\includegraphics[width=0.185\textwidth, cfbox={black} 0.5pt 0pt]{figures/vis/groundtruth_\i.png}}
            \ifnum\i<5
                \hfill
            \fi
        }
    \end{minipage}
    \caption{Visual representation of the model's decision-making process in diverse situations. Rows, from top to bottom, depict: (1) Choices determined by category score, (2) Choices determined by spatial score, (3) Our model's final selection after combining both scores, and (4) The established ground truth. Columns from left to right showcase varying scenarios. The \color{black}green \color{black}bounding box refers to the chosen object, and the \color{black}red \color{black}bounding box refers to the unchosen distractors.}
    \label{fig:real_examples}
\end{figure*}

\begin{table}
\centering
\begin{tabular}{ l | c } 
\hline
\rowcolor[gray]{0.95}
 Object Encoder & Accuracy \\ [0.5ex] 
\hline
PointNet++ & 63.8\% \\  
\hline
PointNet++ \& GCN & 65.5\% \\  
\hline
PointNet++ \& Self-Attention Based & {\bf 70.8\%} \\  
\hline
\end{tabular}
\caption{Comparison of object encoding strategies, presenting the object recognition accuracy achieved with different object encoding techniques. Showing the effectiveness of scene-aware object encoder.}
\label{table:obj_enc}
\vspace{-5mm}
\end{table}

\subsection{Ablation Studies}
\noindent{\bf Effectiveness of Spatial Module:}\\
In \Cref{table:spatialinfo} we conduct a detailed ablation study on the spatial module. Removing the spatial encoder completely and use MLP for direct object location encoding significantly decreased accuracy to 45.0\%. Removing the feedforward layer led to a 20\% reduction in GPU memory usage, but caused a 2\% accuracy drop to 62.4\%. Omitting text-spatial fusion from our model caused a 1.8\% decrease in overall performance, bringing it to 62.6\%.

\noindent{\bf Effectiveness of Scene-Aware Object Encoder:}\\
\Cref{table:obj_enc} demonstrates how incorporating scene context enhances the performance of our object encoder, as compared to a standard PointNet++ encoder \cite{qi2017a}. We introduced a scene-aware module after the PointNet++ layer in two variants: one employing Graph Convolutional Networks (GCN) \cite{kipf2017} and the other using a self-attention mechanism \cite{vaswani2017}. 

For the GCN-based approach, we utilized Euclidean distance in the 3D space to determine neighborhood relations, specifically selecting the 10 nearest neighbors as the basis for graph construction. Our results show significant improvements in object classification accuracy. While the standard PointNet++ encoder achieved an accuracy of 63.8\%, the GCN-based scene-aware encoder increased accuracy to 65.5\%.

The self-attention-based scene-aware encoder further enhanced accuracy to 70.8\%, showing the best performance. This highlights the effectiveness of scene-aware modules in improving object recognition by utilizing information about nearby objects. \\

\noindent{\bf Ablation Study on Spatial Features Aggregation:} \\
\Cref{table:fusion_module} shows our proposed method excels particularly in view-dependent scenarios. We substituted the attention-based aggregate layer with a simple mean function (\RomanNumeralCaps{1}). This change led to a significant drop in accuracy, down to 33.9\%. In contrast, employing max pooling (\RomanNumeralCaps{2}) achieved a 61.7\% accurac. It is worth noting that  without the attention mechanism, accuracy in view-dependent scenarios falls significantly with mean and max pooling.

\subsection{Multi-Modal Prediction} 
In \cref{fig:real_examples}, we illustrate MiKASA's decision-making in various scenarios. (a) shows accurate predictions where category and spatial scores align. (b) highlights accurate object identification with spatial relation challenges due to nearby objects with similar spatial features. (c) depicts scenarios where the model excels in spatial discernment, which is crucial in situations with multiple objects of the same category. (d) presents challenges in predictions where spatial cues are minimal, exemplified by two boxes near a door at a similar distance. Finally, (e) reveals cases where accurate spatial scoring is offset by inadequate category identification, such as with a poorly represented roll of toilet paper. The figure shows our model's decision-making is more explainable and facilitates easier diagnosis of errors. See the supplementary materials for more analysis.
\begin{table}[H]
\vspace{-5mm}
\captionsetup{skip=3pt}
\centering
\footnotesize{
\setlength\tabcolsep{3pt} 
\begin{tabular}{c| l | ccccc } 
\hline
\rowcolor[gray]{0.95}
  & & Overall1 & Easy & Hard & VD & VI \\ [0.5ex] 
\hline
\RomanNumeralCaps{1} & mean & 33.9 & 42.7\% & 25.4\% & 32.8\% & 34.4\%  \\ 
& & \color{mydarkred}$\downarrow$30.5\% & \color{mydarkred}$\downarrow$27.0\% & \color{mydarkred}{\bf $\downarrow$34.0\%} & \color{mydarkred}{\bf $\downarrow$32.6\%} & \color{mydarkred}$\downarrow$29.6\%\\
\hline
\RomanNumeralCaps{2} & max & 61.4\% & 67.8\% & 55.2\% & 60.6\% & 61.7\% \\ 
& pool & \color{mydarkred}$\downarrow$3.0\% & \color{mydarkred}$\downarrow$1.9\% & \color{mydarkred}{\bf $\downarrow$4.2\%} & \color{mydarkred}{\bf $\downarrow$ 4.8\%} & \color{mydarkred}$\downarrow$2.3\% \\
\hline
&  \bf{Ours} & \bf{64.4\%} & \bf{69.7\%} & \bf{59.4\%} & \bf{65.4\%} & \bf{64.0\%} \\  
\hline
\end{tabular}
}
\caption{Ablations on fusion module}
\label{table:fusion_module}
\end{table}
\vspace{-5mm}

\section{Conclusion}
\label{sec:conclusion}
In our study, we introduced the MiKASA (Multi-Key-Anchor Scene-Aware) Transformer, an innovative model designed to address the challenges in 3D visual grounding. This model uniquely combines a scene-aware object encoder with a multi-key-anchor technique, significantly enhancing object recognition and spatial understanding in 3D environments. The scene-aware object encoder effectively tackles object categorization issues, while the multi-key-anchor technique offers improved interpretation of spatial relationships and viewpoints. The results demonstrate that MiKASA outperforms current state-of-the-art models in both accuracy and explainability, underscoring its efficacy in advancing 3D visual grounding research. For future work, we suggest enhancing the model to explicitly preserve the directional information of objects post-encoding, aiming to further refine accuracy in view-dependent scenarios.

\noindent{\bf Acknowledgement:}
This research has been partially funded by EU project FLUENTLY (GA: Nr 101058680) and the BMBF project SocialWear (01IW20002).
\newpage
{\small
\bibliographystyle{ieee_fullname}
\bibliography{egbib}
}
\end{document}